\documentclass{article}

\usepackage{arxiv}

\usepackage[utf8]{inputenc} 
\usepackage[T1]{fontenc}    
\usepackage{lmodern}        
\usepackage{hyperref}       
\usepackage{url}            
\usepackage{booktabs}       
\usepackage{amsfonts}       
\usepackage{nicefrac}       
\usepackage{microtype}      
\usepackage{graphicx}

\title{Explainable Boosting Machines with Sparsity - Maintaining Explainability in High-Dimensional Settings}

\author{
    Brandon M. Greenwell
    \thanks{Use footnote for providing further information about author (webpage, alternative address)---\emph{not} for acknowledging funding agencies. Optional.}
   \\
    84.51° / University of Cincinnati \\
  Cincinnati \\
  \texttt{\href{mailto:brandon.greenwell@8451.com}{\nolinkurl{brandon.greenwell@8451.com}}} \\
   \And
    Annika Dahlmann
   \\
    84.51° \\
  Chicago \\
  \texttt{\href{mailto:annika.dahlmann@8451.com}{\nolinkurl{annika.dahlmann@8451.com}}} \\
   \And
    Saurabh Dhoble
   \\
    84.51° \\
  Cincinnati \\
  \texttt{\href{mailto:saurabh.dhoble@8451.com}{\nolinkurl{saurabh.dhoble@8451.com}}} \\
  }

\usepackage{color}
\usepackage{fancyvrb}

\DefineVerbatimEnvironment{Highlighting}{Verbatim}{commandchars=\\\{\}}
\usepackage{framed}
\definecolor{shadecolor}{RGB}{248,248,248}
\newenvironment{Shaded}{\begin{snugshade}}{\end{snugshade}}

\newcommand{\AttributeTok}[1]{\textcolor[rgb]{0.77,0.63,0.00}{#1}}

\newcommand{\BuiltInTok}[1]{#1}

\newcommand{\CommentTok}[1]{\textcolor[rgb]{0.56,0.35,0.01}{\textit{#1}}}

\newcommand{\ConstantTok}[1]{\textcolor[rgb]{0.00,0.00,0.00}{#1}}
\newcommand{\ControlFlowTok}[1]{\textcolor[rgb]{0.13,0.29,0.53}{\textbf{#1}}}

\newcommand{\DecValTok}[1]{\textcolor[rgb]{0.00,0.00,0.81}{#1}}

\newcommand{\FloatTok}[1]{\textcolor[rgb]{0.00,0.00,0.81}{#1}}
\newcommand{\FunctionTok}[1]{\textcolor[rgb]{0.00,0.00,0.00}{#1}}
\newcommand{\ImportTok}[1]{#1}

\newcommand{\KeywordTok}[1]{\textcolor[rgb]{0.13,0.29,0.53}{\textbf{#1}}}
\newcommand{\NormalTok}[1]{#1}
\newcommand{\OperatorTok}[1]{\textcolor[rgb]{0.81,0.36,0.00}{\textbf{#1}}}
\newcommand{\OtherTok}[1]{\textcolor[rgb]{0.56,0.35,0.01}{#1}}

\newcommand{\SpecialCharTok}[1]{\textcolor[rgb]{0.00,0.00,0.00}{#1}}

\newcommand{\StringTok}[1]{\textcolor[rgb]{0.31,0.60,0.02}{#1}}
\newcommand{\VariableTok}[1]{\textcolor[rgb]{0.00,0.00,0.00}{#1}}

\providecommand{\tightlist}{%
  \setlength{\itemsep}{0pt}\setlength{\parskip}{0pt}}

\usepackage{longtable,booktabs,array}
\usepackage{calc} 
\usepackage{etoolbox}
\makeatletter
\patchcmd\longtable{\par}{\if@noskipsec\mbox{}\fi\par}{}{}
\makeatother
\IfFileExists{footnotehyper.sty}{\usepackage{footnotehyper}}{\usepackage{footnote}}
\makesavenoteenv{longtable}

\newlength{\cslhangindent}
\newlength{\csllabelwidth}
\newlength{\cslentryspacingunit} 
  {}%
  {\par}
\newenvironment{CSLReferences}[2] 
 {
  \setlength{\parindent}{0pt}
  \ifodd #1
  \let\oldpar\par
  \def\par{\hangindent=\cslhangindent\oldpar}
  \fi
  \setlength{\parskip}{#2\cslentryspacingunit}
 }%
 {}
\usepackage{calc}

\usepackage{algorithmic}
\usepackage{hyperref}
\usepackage{float}
\usepackage{amsmath}

\hypersetup{colorlinks=true,linkcolor=blue,urlcolor=magenta}
\begin{document}
\maketitle

\begin{abstract}
Compared to ``black-box'' models, like random forests and deep neural networks, explainable boosting machines (EBMs) are considered ``glass-box'' models that can be competitively accurate while also maintaining a higher degree of transparency and explainability. However, EBMs become readily less transparent and harder to interpret in high-dimensional settings with many predictor variables; they also become more difficult to use in production due to increases in scoring time. We propose a simple solution based on the \emph{least absolute shrinkage and selection operator} (LASSO) that can help introduce sparsity by reweighting the individual model terms and removing the less relevant ones, thereby allowing these models to maintain their transparency and relatively fast scoring times in higher-dimensional settings. In short, post-processing a fitted EBM with many (i.e., possibly hundreds or thousands) of terms using the LASSO can help reduce the model's complexity and drastically improve scoring time. We illustrate the basic idea using two real-world examples with code.
\end{abstract}


\floatstyle{ruled}
\newfloat{algo}{thp}{lop}
\floatname{algo}{Algorithm}

\hypertarget{introduction}{%
\section{Introduction}\label{introduction}}

Explainable boosting machines (Nori et al. 2019), or EBMs for short, are a modern class of \emph{generalized additive models} (GAMs) that can offer both competitive accuracy and explicit transparency and explainability, which are often considered to be two opposing goals of a machine learning (ML) model. For example, full-complexity models, like random forests (Breiman 2001), tend to be highly competitive in terms of accuracy, but are readily less transparent and explainable (due in part to the high-order interaction effects often captured by such black-box models). Essentially, with EBMs, it's possible to ``have your cake and eat it too.''

In short, EBM models have the general form

\[
g\left(E\left[y\right]\right) = \theta_0 + \sum f_i\left(x_i\right) + \sum f_{ij}\left(x_i, x_j\right),
\]
where

\begin{itemize}
\tightlist
\item
  \(g\) is a \emph{link function} that allows the model to handle various response types (e.g., the \emph{logit} link for logistic regression or the identify function for ordinary regression with continuous outcomes);
\item
  \(\theta_0\) is a constant intercept (or bias term);
\item
  \(f_i\) is the \emph{term contribution} (or \emph{shape function}) for predictor \(x_i\) (i.e., it captures the main effect of \(x_i\));
\item
  \(f_{ij}\) is the term contribution for the pair of predictors \(x_i\) and \(x_j\) (i.e., it captures the joint effect, or pairwise interaction effect of \(x_i\) and \(x_j\)).
\end{itemize}

Similar to \emph{generalized additive models plus interactions} (Lou et al. 2013), or GA\textsuperscript{2}Ms for short, the pairwise interaction terms are determined automatically using the FAST algorithm described in Lou et al. (2013). In short, FAST is a novel, computationally efficient method for ranking all possible pairs of feature candidates for inclusion into the model (by default, the top 10 pairwise interactions are used). The primary difference between EBMs and GA\textsuperscript{2}Ms is in how the shape functions are estimated. In particular, EBMs use \emph{cyclic gradient boosting} (Nori et al. 2019; Wick, Kerzel, and Feindt 2020) to estimate the shape functions for each feature (and selected pairs of interactions) in a round-robin fashion using a low learning rate to help ensure that the order in which the feature effects are estimated does not matter. Estimating each feature one-at-a-time in a round-robin fashion also helps mitigate potential issues with \emph{collinearity} between predictors (Nori et al. 2019; Wick, Kerzel, and Feindt 2020).

However, in contrast to the more common gradient boosting machine (J. H. Friedman 2001, 2002), or GBM for short, which can ignore irrelevant inputs, EBMs include at least one term in the model for each feature: one main effect (\(f_i\)) for each predictor, and a term for each selected pairwise interaction effect. This is due to the cyclic nature of the underlying boosting framework. For example, an EBM applied to a training set with \(p = 300\) features will result in a model with at least 300 terms.

While EBMs are considered \emph{glass-box} models, an EBM with, say, hundreds or thousands of terms, starts to become much less transparent and explainable. Moreover, the larger the fitted model (i.e., the more terms there are), then the more time it will take the EBM to make predictions, making larger models less fit for deployment and production.

To this end, we propose in Section \ref{sec:isle} a way to introduce sparsity into a fitted EBM by rescaling the individual terms (i.e., the \(f_i\) and \(f_{ij}\)) via regression coefficients estimated from a method called the LASSO (Tibshirani 1996). Consequently, by nature of the \(L_1\) regularization enforced by the LASSO, many of these coefficients can be estimated to be zero, resulting in a reduced EBM with far fewer terms (and hopefully, comparable accuracy)!

\hypertarget{post-processing-ebms-with-the-lasso}{%
\section{Post-processing EBMs with the LASSO}\label{post-processing-ebms-with-the-lasso}}

\label{sec:isle}

Tree-based ensembles, like random forests and GBMs, often do a good job in building a prediction model for tabular data. However, these ensembles often involve hundreds or thousands of trees which can limit their use in production since more trees requires more memory and increases prediction time.

To help overcome these issues, J. H. Friedman and Popescu (2003) introduced the concept of \emph{importance sampled learning ensembles} (ISLEs). The main point here is that large tree ensembles can sometimes be simplified by post-processing them with the LASSO. Such post-processing can often maintain or, in some cases, improve the accuracy of the original ensemble while dramatically reducing the number of trees (thereby resulting in lower memory requirements and faster prediction times). For full details, see J. H. Friedman and Popescu (2003), Hastie, Tibshirani, and Friedman (2009, sec. 16.3.1), and Efron and Hastie (2016, 346--47).

The idea is to reweight each tree's contribution to the overall prediction using regression coefficients estimated with the LASSO. Since the LASSO uses an \(L_1\) penalty, many of the resulting coefficients can be shrunk all the way to zero, resulting in an ensemble with far fewer trees and (hopefully) comparable, if not better accuracy. This is important to consider in real applications since tree ensembles can sometimes require thousands of decision trees to reach peak performance, often resulting in a large model to maintain in memory and slower scoring times (aspects that are important to consider before deploying a model in a production process).

For tree-based ensembles, the LASSO-based post-processing procedure essentially involves fitting an \(L_1\)-penalized regression model of the form

\[
\nonumber
  \mathop{\mathrm{minimize}}_{\left\{\beta_b\right\}_{b = 1}^B} \sum_{i = 1}^N L\left[y_i, \sum_{b = 1}^B \widehat{f}_b\left(\boldsymbol{x}_i\right) \beta_b\right] + \lambda\sum_{b = 1}^B \left\lvert \beta_b \right\rvert,
\]

where \(L\) is a suitable loss function (e.g., residual sum of squares for regression or binomial deviance for binary outcomes), \(\widehat{f}_b\left(\boldsymbol{x}_i\right)\) (\(b = 1, 2, \dots, B\)) is the prediction from the \(b\)-th tree for observation \(\boldsymbol{x}_i\), \(\beta_b\) are fixed, but unknown coefficients to be estimated via the LASSO, and \(\lambda\) is the \(L_1\) penalty to be applied (and estimated using an independent test sample, as briefly discussed later).

The wonderful and efficient \href{https://cran.r-project.org/package=glmnet}{glmnet} package (J. Friedman et al. 2021) for R can be used to fit the entire LASSO regularization path\footnote{The \textbf{glmnet} package actually implements the entire \textit{elastic-net} regularization path for many types of generalized linear models; the LASSO is just a special case of the elastic net, which combines both the LASSO and ridge (i.e., $L_2$) penalties.}; that is, computes the estimated coefficients for a grid of relevant \(\lambda\) values, which can be chosen via cross-validation, or an independent test set.

We propose that the same idea can successfully be applied to EBMs in an effort to reduce model complexity in higher-dimensional settings, allowing EBMs to retain their glass-box nature and reduce scoring time. Note that the number of trees in a tree-based ensemble, like a random forest, has no impact on transparency and explainability. For instance, a random forest with 3 trees is just as black-box as a random forest with 3000 trees. So, post-processing such tree-based ensembles is mostly concerned with reducing model size and improving scoring time. In contrast, treating the individual terms in an EBM in an analogous manner has ramifications on model size, scoring time, transparency, and explainability!

To illustrate, suppose we used an EBM regressor to fit the following additive regression model (i.e., using the identity link):

\[
E\left(y\right) = \theta_0 + \sum_{i=1}^p f_i\left(x_i\right),
\]

where \(\theta_0\) is a constant intercept (or bias) term and the shape functions, \(f_i\), are estimated via cyclic gradient boosting in the case of an EBM. We propose to use \(L_1\) regularization to rescale the the individual term contributions which, in many cases, can result in several of the original terms being dropped from the model (i.e., their estimated coefficient shrinks all the way to zero).

For the regression case, we propose reweighting the individual term contributions (i.e., the shape functions \(f_i\)) by solving the following LASSO problem\footnote{For brevity, we're ignoring pairwise interaction terms.}:

\begin{equation}
\mathop{\mathrm{minimize}}_{{\left\{\beta_b\right\}}_{b=0}^p} \sum_{i = 1}^n \left(y_i - \beta_0 - \sum_{j=1}^p\beta_j f_j\left(x_{ij}\right)\right)^2 + \lambda \sum_{j=1}^p \lvert \beta_j \rvert
\label{eq:ebm-isle}
\end{equation}

The original intercept, \(\theta_0\), can be ignored and re-estimated through the LASSO solution (i.e., \(\beta_0\)), or it can be treated as an \emph{offset} term in the LASSO model (we haven't seen much of a difference between either approach in our own experiments).

Given the estimated LASSO coefficients from solving \ref{eq:ebm-isle}, the final, post-processed EBM will have the form

\[
E\left(y\right) = \beta_0 + \sum_{i=1}^p \beta_i f_i\left(x_i\right),
\]

where, ideally, many of the estimated term weights \(\left\{\beta_i\right\}_{i=1}^p\) will be zero, meaning those terms can be dropped entirely from the model. The general steps for post-processing a fitted EBM with the LASSO are outlined in Algorithm 1.

\begin{algo}[!htb]
Given training and test data sets, $\left\{\boldsymbol{X}_{trn}, \boldsymbol{y}_{trn}\right\}$ and $\left\{\boldsymbol{X}_{tst}, \boldsymbol{y}_{tst}\right\}$, respectively, do the following:
\begin{enumerate}

  \item Fit an EBM to the training data to obtain the initial fit $$f_0\left(\boldsymbol{x}\right) = \theta_0 + \sum_{i}f_i\left(x_i\right) + \sum_{ij}f_{ij}\left(x_i, x_j\right).$$ 
  
  \begin{itemize}
    \item Ideally, the number of predictors (i.e., number of columns in $\boldsymbol{X}_{trn}$) should be large (e.g., in the hundreds or thousands).
    \item Be sure to use either cross-validation or a third, independent validation sample for tuning and/or \textit{early stopping} (i.e., do not use the test data defined above in this step or you'll introduce leakage into the process). 
    \item Define $p$ to be the total number of terms in $f_0$ (i.e., number of main effects plus number of pairwise interactions).
  \end{itemize}

  \item Create a new matrix of training features, denoted $\boldsymbol{X}_{trn}^\star$, where the $i$-th column refers to the $i$-th term contribution of $f_0$; for example, if the first column of $\boldsymbol{X}_{trn}$ represents the values of predictor $x_1$, then first column of $\boldsymbol{X}_{trn}^\star$ corresponds to the values of $f_1\left(x_1\right)$. In other words, you can think of the columns of $\boldsymbol{X}_{trn}^\star$ as a set of newly engineered features determined by the initial EBM fit. Do the same for the predictor values in the test set as well, and call the resulting matrix $\boldsymbol{X}_{tst}^\star$.
  
  \item Use the modified training data, $\left\{\boldsymbol{X}_{trn}^\star, \boldsymbol{y}_{trn}\right\}$, to fit the entire LASSO regularization path.
  
  \item Use the modified test data, $\left\{\boldsymbol{X}_{tst}^\star, \boldsymbol{y}_{tst}\right\}$, to select the estimated coefficients corresponding to the optimal value of $\lambda$ along the regularization path from the previous step; for simplicity, let $\beta_0$, $\left\{\beta_i\right\}$, and $\left\{\beta_{ij}\right\}$ represent the selected intercept and sets of coefficients for the main effects and pairwise interactions, respectively, for a total of $p + 1$ coefficients. 
  
  \item Edit the initial EBM, $f_0$, by multiplying each term contributions by its associated LASSO coefficient to obtain the post-processed EBM $$f\left(\boldsymbol{x}\right) = \beta_0 + \sum_{i}\beta_if_i\left(x_i\right) + \sum_{ij}\beta_{ij}f_{ij}\left(x_i, x_j\right).$$
  
  \begin{itemize}
    \item If any of the estimated coefficients are zero, then update the model by removing those terms (this can be done in available EBM software, and the relevant code to accomplish this is illustrated in the ALS example in Section \ref{sec:als}.)
  \end{itemize}
  
  \item Return the post-processed EBM, $f\left(\boldsymbol{x}\right)$.
  
  \item (Optional, but recommended) Validate the post-processed EBM on a final, separate holdout set.

\end{enumerate}
\caption{Steps for post-processing a fitted EBM with the LASSO.}
\end{algo}

Note that this approach can be applied to any link function, and also allow for pairwise interaction terms. An example using the logit to model a binary outcome (e.g., as in logistic regression) is given in Section \ref{sec:coil}. Further, it seems intuitive to also force the LASSO coefficients be strictly non-negative; a negative coefficient would cause a term to contribute in the opposite direction of the original model, which seems unreasonable. This approach is referred to as the \emph{non-negative LASSO} (Wu, Yang, and Liu 2014). Forcing the estimated coefficients to be non-negative along the entire regularization path is possible with most LASSO modeling software (e.g., \href{https://cran.r-project.org/package=glmnet}{glmnet} in R and \href{https://scikit-learn.org/stable/modules/classes.html\#module-sklearn.linear_model}{sklearn}'s \texttt{Lasso} submodule in Python).

\hypertarget{examples-with-code}{%
\section{Examples with code}\label{examples-with-code}}

We now provide two examples from real applications of ML: one for regression and one for binary classification. We'll start with the regression example. For both examples, we make use of the open-source Python library \href{https://interpret.ml/}{InterpretML} (Nori et al. 2019). Note that the following examples require version 0.4.4 to take advantage of the new \texttt{scale()} and \texttt{sweep()} methods for model editing that were added in \href{https://github.com/interpretml/interpret/releases}{this release}.

\hypertarget{predicting-als-progression-the-dream-challenge-prediction-prize}{%
\subsection{Predicting ALS progression (the DREAM challenge prediction prize)}\label{predicting-als-progression-the-dream-challenge-prediction-prize}}

\label{sec:als}

In this section, we'll look at a brief example using the ALS data from Efron and Hastie (2016, 349). A description of the data, along with the original source and download instructions, can be found at \url{https://hastie.su.domains/CASI/data.html}.

These data concern 1822 observations on \emph{amyotrophic lateral sclerosis} (ALS or Lou Gehrig's disease) patients. The goal is to predict ALS progression over time, as measured by the slope (or derivative) of a functional rating score (i.e., column \texttt{dFRS}), using 369 available predictors obtained from patient visits. The data were originally part of the DREAM-Phil Bowen ALS Predictions Prize4Life challenge. The winning solution (Küffner et al. 2015) used a Bayesian tree ensemble quite similar to a random forest, while Efron and Hastie (2016, chap. 17) analyzed the data using GBMs; the latter used same ISLE idea to show how a GBM model can be simplified while maintaining accuracy close to the original fit.

In the code chunk below, we'll import the necessary libraries and read in the ALS data:

\begin{Shaded}
\begin{Highlighting}[]
\ImportTok{import}\NormalTok{ numpy }\ImportTok{as}\NormalTok{ np}
\ImportTok{import}\NormalTok{ pandas }\ImportTok{as}\NormalTok{ pd}

\ImportTok{from}\NormalTok{ interpret.glassbox }\ImportTok{import}\NormalTok{ ExplainableBoostingRegressor }\ImportTok{as}\NormalTok{ EBR}
\ImportTok{from}\NormalTok{ sklearn.metrics }\ImportTok{import}\NormalTok{ mean\_squared\_error}
\ImportTok{from}\NormalTok{ sklearn.linear\_model }\ImportTok{import}\NormalTok{ Lasso, lasso\_path}


\CommentTok{\# Read in ALS data (analysis adapted from "Computer Age Statistical Inference")}
\NormalTok{url }\OperatorTok{=} \StringTok{\textquotesingle{}https://hastie.su.domains/CASI\_files/DATA/ALS.txt\textquotesingle{}}
\NormalTok{als }\OperatorTok{=}\NormalTok{ pd.read\_csv(url, delim\_whitespace}\OperatorTok{=}\VariableTok{True}\NormalTok{)}
\NormalTok{als.head()  }\CommentTok{\# inspect first few rows}
\end{Highlighting}
\end{Shaded}

\begin{verbatim}
##    testset      dFRS  ...  sd.slope.bp.systolic  slope.bp.systolic.slope
## 0     True -0.915388  ...              4.163843                 5.003010
## 1     True -0.107931  ...              4.909140                -8.920469
## 2     True -0.557448  ...              6.028041                14.677878
## 3     True -0.296461  ...             42.048818                41.458842
## 4     True -1.087024  ...              7.321307                 3.759702
## 
## [5 rows x 371 columns]
\end{verbatim}

Next, we'll split the data into train and test sets using the provided test set indicator (i.e., column \texttt{testset}) and fit a simple EBM; for details on the available parameters, see associated API documentation at \url{https://interpret.ml/docs/ebm.html\#api}. As suggested in the EBM documentation\footnote{See the section on parameter tuning at \url{https://interpret.ml/docs/faq.html}.}, we simply increased the number of inner and outer bags to 25 in order to improve accuracy and provide a smoother fit.

\begin{Shaded}
\begin{Highlighting}[]
\CommentTok{\# Split into train/test (and drop test set indicator)}
\NormalTok{als\_trn }\OperatorTok{=}\NormalTok{ als[als[}\StringTok{\textquotesingle{}testset\textquotesingle{}}\NormalTok{] }\OperatorTok{==} \VariableTok{False}\NormalTok{].drop(}\StringTok{\textquotesingle{}testset\textquotesingle{}}\NormalTok{, axis}\OperatorTok{=}\DecValTok{1}\NormalTok{)}
\NormalTok{als\_tst }\OperatorTok{=}\NormalTok{ als[als[}\StringTok{\textquotesingle{}testset\textquotesingle{}}\NormalTok{] }\OperatorTok{==} \VariableTok{True}\NormalTok{].drop(}\StringTok{\textquotesingle{}testset\textquotesingle{}}\NormalTok{, axis}\OperatorTok{=}\DecValTok{1}\NormalTok{)}

\CommentTok{\# Separate predictors from response}
\NormalTok{X\_trn }\OperatorTok{=}\NormalTok{ als\_trn.drop(}\StringTok{\textquotesingle{}dFRS\textquotesingle{}}\NormalTok{, axis}\OperatorTok{=}\DecValTok{1}\NormalTok{)}
\NormalTok{X\_tst }\OperatorTok{=}\NormalTok{ als\_tst.drop(}\StringTok{\textquotesingle{}dFRS\textquotesingle{}}\NormalTok{, axis}\OperatorTok{=}\DecValTok{1}\NormalTok{)}
\NormalTok{y\_trn }\OperatorTok{=}\NormalTok{ als\_trn[}\StringTok{\textquotesingle{}dFRS\textquotesingle{}}\NormalTok{]}
\NormalTok{y\_tst }\OperatorTok{=}\NormalTok{ als\_tst[}\StringTok{\textquotesingle{}dFRS\textquotesingle{}}\NormalTok{]}

\CommentTok{\# Fit an EBM}
\NormalTok{ebm }\OperatorTok{=}\NormalTok{ EBR(inner\_bags}\OperatorTok{=}\DecValTok{25}\NormalTok{, outer\_bags}\OperatorTok{=}\DecValTok{25}\NormalTok{)}
\NormalTok{ebm.fit(X\_trn, y}\OperatorTok{=}\NormalTok{y\_trn)}
\end{Highlighting}
\end{Shaded}

\begin{verbatim}
## ExplainableBoostingRegressor(inner_bags=25, outer_bags=25)
\end{verbatim}

The fitted EBM model contains 379 terms (369 main effect terms, one for each predictor, and 10 pairwise interaction terms) plus an intercept. The mean square error for this model on the test data was 0.265. The top 20 terms, as measured by the mean absolute contribution to the model's predictions (e.g., \(avg\left\{\lvert f_i\left(x_i\right) \rvert\right\}_{i=1}^n\)), are displayed in Figure \ref{fig:ebm-als-vi} below. Here, we see that \texttt{Onset.Delta} is an important predictor through both a main effect and various pairwise interactions with other features (e.g., \texttt{max.dressing}).

\begin{figure}[!htb]

{\centering \includegraphics[width=1\linewidth]{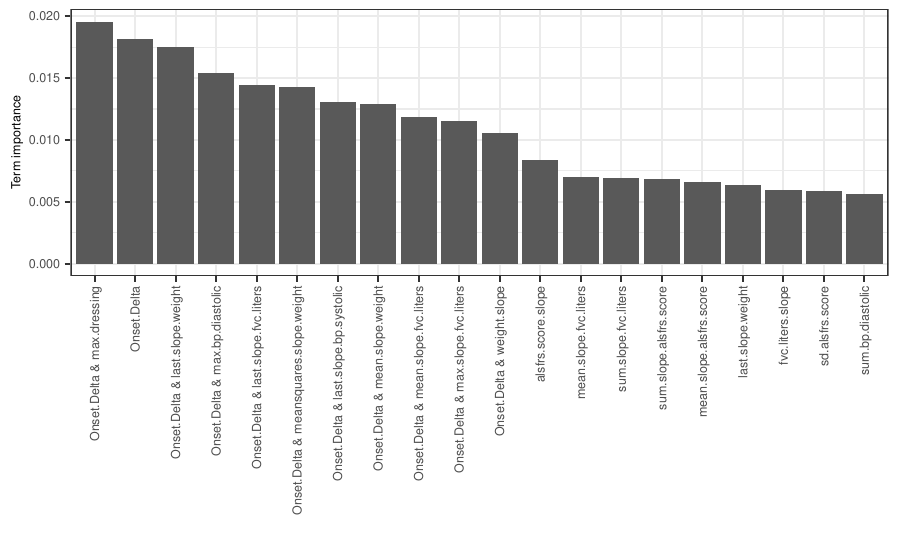} 

}

\caption{Term importance scores for the top 20 terms, as measured by the mean absolute contribution of each term to the model's predictions.}\label{fig:ebm-als-vi}
\end{figure}

\begin{figure}[!htb]

{\centering \includegraphics[width=1\linewidth]{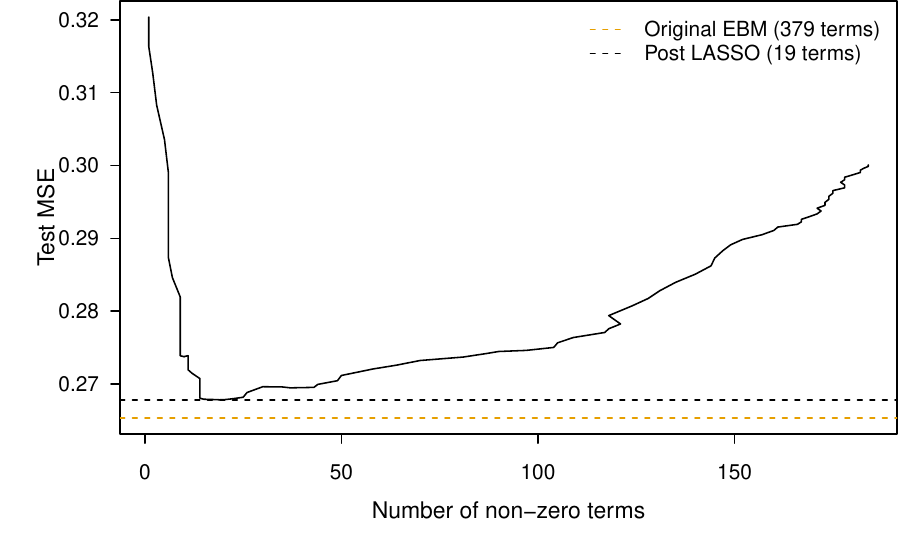} 

}

\caption{Post-processing of the terms produced by applying an EBM to the ALS data. Shown is the test MSE as a function of the number of terms (i.e., non-zero coefficients) selected by the (nonnegative) LASSO (not counting the intercept). In this case, the LASSO post-processed EBM does not improve the prediction error of the original model (nor does it increase it by much), but it did provide a 95\% reduction in the number of terms, going from the 379 down to 19!}\label{fig:als-ebm-lasso-plot}
\end{figure}

Next, we construct new train and test data sets based on the individual term contribution from the EBM model. For instance, the new training set will use \(f_1\left(x_1\right)\) in place of \(x_1\), as well as the pairwise contribution terms (i.e., we're treating the terms from the initial model as the new inputs). We then fit the LASSO regularization path and compute the test MSE as a function of the number of terms in the model (i.e., non-zero coefficients); it's important to use an independent test set here to avoid \emph{data leakage} (Kapoor and Narayanan 2023). The results are displayed in Figure \ref{fig:als-ebm-lasso-plot}. (Note that the LASSO regularization parameter, \(\lambda\), is referred to as \texttt{alpha} in \textbf{sklearn}.)

\begin{Shaded}
\begin{Highlighting}[]
\CommentTok{\# Convert train/test sets into their associated term contributions}
\NormalTok{X\_trn\_tc }\OperatorTok{=}\NormalTok{ ebm.predict\_and\_contrib(X\_trn)[}\DecValTok{1}\NormalTok{]}
\NormalTok{X\_tst\_tc }\OperatorTok{=}\NormalTok{ ebm.predict\_and\_contrib(X\_tst)[}\DecValTok{1}\NormalTok{]}

\CommentTok{\# Fit LASSO path and grab alpha values}
\NormalTok{ebm\_lasso\_path }\OperatorTok{=}\NormalTok{ lasso\_path(X\_trn\_tc, y}\OperatorTok{=}\NormalTok{y\_trn, positive}\OperatorTok{=}\VariableTok{True}\NormalTok{)}
\NormalTok{alphas }\OperatorTok{=}\NormalTok{ ebm\_lasso\_path[}\DecValTok{0}\NormalTok{]}

\CommentTok{\# Fit a LASSO for each alpha; compute test MSE and number of non{-}zero coefs}
\NormalTok{mses }\OperatorTok{=}\NormalTok{ []}
\NormalTok{nterms }\OperatorTok{=}\NormalTok{ []}
\ControlFlowTok{for}\NormalTok{ alpha }\KeywordTok{in}\NormalTok{ alphas:}
\NormalTok{    \_lasso }\OperatorTok{=}\NormalTok{ Lasso(alpha}\OperatorTok{=}\NormalTok{alpha, positive}\OperatorTok{=}\VariableTok{True}\NormalTok{)}
\NormalTok{    \_lasso.fit(X\_trn\_tc, y}\OperatorTok{=}\NormalTok{y\_trn)}
\NormalTok{    \_mse }\OperatorTok{=}\NormalTok{ mean\_squared\_error(y\_tst, y\_pred}\OperatorTok{=}\NormalTok{\_lasso.predict(X\_tst\_tc))}
\NormalTok{    mses.append(\_mse)}
\NormalTok{    nterms.append(np.count\_nonzero(\_lasso.coef\_))}
\end{Highlighting}
\end{Shaded}

The final LASSO model (i.e., based on the optimal \(\lambda\) associated with the smallest test MSE) is given below. Note that the LASSO estimated a new intercept parameter, which we'll use to update the intercept attribute of the EBM model we fit earlier.

\begin{Shaded}
\begin{Highlighting}[]
\CommentTok{\# Fit a LASSO using the "best" alpha }
\NormalTok{alpha\_best }\OperatorTok{=}\NormalTok{ alphas[np.argmin(mses)]}
\NormalTok{ebm\_lasso }\OperatorTok{=}\NormalTok{ Lasso(alpha}\OperatorTok{=}\NormalTok{alpha\_best, positive}\OperatorTok{=}\VariableTok{True}\NormalTok{)}
\NormalTok{ebm\_lasso.fit(X\_trn\_tc, y}\OperatorTok{=}\NormalTok{y\_trn)}
\end{Highlighting}
\end{Shaded}

\begin{verbatim}
## Lasso(alpha=0.0003722658525823356, positive=True)
\end{verbatim}

\begin{Shaded}
\begin{Highlighting}[]
\NormalTok{np.count\_nonzero(ebm\_lasso.coef\_)  }\CommentTok{\# number of non{-}zero coefficients}
\end{Highlighting}
\end{Shaded}

\begin{verbatim}
## 19
\end{verbatim}

\begin{Shaded}
\begin{Highlighting}[]
\CommentTok{\# Compare intercepts}
\NormalTok{ebm.intercept\_, ebm\_lasso.intercept\_}
\end{Highlighting}
\end{Shaded}

\begin{verbatim}
## (-0.6799731167196948, -0.6802357522044189)
\end{verbatim}

Finally, after obtaining the LASSO coefficients corresponding to the optimal values of the regularization parameter \(\lambda\), we can edit the original EBM model by calling the new \texttt{scale()} and \texttt{sweep()} methods. This is demonstrated in the following code snippet. Note that \texttt{scale()} works on individual terms and takes two arguments, \texttt{term} and \texttt{factor}, that allows you to scale the individual term contribution (i.e., \texttt{term}) by a constant factor (i.e., \texttt{factor}). This is where we can pass in the corresponding LASSO coefficients, which we do below via a simple for loop. (Note that we have to manually adjust the intercept attribute.)

\begin{Shaded}
\begin{Highlighting}[]
\CommentTok{\# Rescale terms using LASSO coefficients }
\ControlFlowTok{for}\NormalTok{ idx, x }\KeywordTok{in} \BuiltInTok{enumerate}\NormalTok{(ebm.term\_names\_):}
\NormalTok{    ebm.scale(idx, factor}\OperatorTok{=}\NormalTok{ebm\_lasso.coef\_[idx])}
\end{Highlighting}
\end{Shaded}

\begin{Shaded}
\begin{Highlighting}[]
\NormalTok{ebm.intercept\_ }\OperatorTok{=} \BuiltInTok{float}\NormalTok{(ebm\_lasso.intercept\_)  }\CommentTok{\# also update intercept}
\end{Highlighting}
\end{Shaded}

This modifies the output to reflect that each term has been rescaled; for instance, the main effect term \(f_i\left(x_i\right)\) is replaced with \(\beta_if_i\left(x_i\right)\). This will impact all the global and local explanations as well. For example, if a specific term has an associated coefficient of zero, then its importance and contribution to the model should also be zero! As a brief sanity check, the code below looks as the number of terms with zero and non-zero importance scores.

\begin{Shaded}
\begin{Highlighting}[]
\CommentTok{\# Check term importance scores}
\NormalTok{term\_imp }\OperatorTok{=}\NormalTok{ ebm.term\_importances()}
\NormalTok{np.count\_nonzero(term\_imp)  }\CommentTok{\# terms with non{-}zero coefficients}
\end{Highlighting}
\end{Shaded}

\begin{verbatim}
## 19
\end{verbatim}

\begin{Shaded}
\begin{Highlighting}[]
\BuiltInTok{len}\NormalTok{(term\_imp) }\OperatorTok{{-}}\NormalTok{ np.count\_nonzero(term\_imp)  }\CommentTok{\# terms with coefficient of zero}
\end{Highlighting}
\end{Shaded}

\begin{verbatim}
## 360
\end{verbatim}

Since many of the terms in the modified EBM now provide zero contribution to the model and its predictions, it would make sense to remove these terms (and their associated components, like term importance scores) from the model. This can be done with the \texttt{sweep()} method, as illustrated below.

\begin{Shaded}
\begin{Highlighting}[]
\CommentTok{\# Remove unused terms (e.g., those with zero importance)}
\NormalTok{ebm.sweep()}
\BuiltInTok{len}\NormalTok{(ebm.term\_names\_)  }\CommentTok{\# sanity check (should only have 19 terms now)}
\end{Highlighting}
\end{Shaded}

\begin{verbatim}
## 19
\end{verbatim}

\begin{Shaded}
\begin{Highlighting}[]
\CommentTok{\# Check term importance scores for reduced EBM (should only have 19 terms not}
\CommentTok{\# counting the intercept)}
\NormalTok{new\_term\_vi }\OperatorTok{=}\NormalTok{ pd.DataFrame(\{}
  \StringTok{\textquotesingle{}term\textquotesingle{}}\NormalTok{ : ebm.term\_names\_, }\StringTok{\textquotesingle{}importance\textquotesingle{}}\NormalTok{ : ebm.term\_importances()}
\NormalTok{\})}
\NormalTok{new\_term\_vi.sort\_values(by}\OperatorTok{=}\NormalTok{[}\StringTok{\textquotesingle{}importance\textquotesingle{}}\NormalTok{], ascending}\OperatorTok{=}\VariableTok{False}\NormalTok{)}
\end{Highlighting}
\end{Shaded}

\begin{verbatim}
##                                       term  importance
## 15         Onset.Delta & last.slope.weight    0.052911
## 1                       alsfrs.score.slope    0.047913
## 9               Onset.Delta & max.dressing    0.039363
## 0                        last.alsfrs.score    0.037614
## 11          Onset.Delta & max.bp.diastolic    0.035902
## 18    Onset.Delta & last.slope.bp.systolic    0.034695
## 13     Onset.Delta & last.slope.fvc.liters    0.031072
## 14     Onset.Delta & mean.slope.fvc.liters    0.030386
## 2                              last.speech    0.026974
## 10              Onset.Delta & weight.slope    0.024253
## 16         Onset.Delta & mean.slope.weight    0.023105
## 17  Onset.Delta & meansquares.slope.weight    0.022325
## 12      Onset.Delta & max.slope.fvc.liters    0.021039
## 5                   min.slope.alsfrs.score    0.008733
## 8                    mean.slope.svc.liters    0.007403
## 3                         fvc.liters.slope    0.002594
## 4                         sum.bp.diastolic    0.002090
## 6                        mean.slope.speech    0.000610
## 7                     max.slope.svc.liters    0.000289
\end{verbatim}

This edited model is now leaner and will score much faster than the original fit since it involves far fewer terms!

\hypertarget{predicting-and-explaining-policy-ownership-the-coil-challenge}{%
\subsection{Predicting and explaining policy ownership (the CoIL challenge)}\label{predicting-and-explaining-policy-ownership-the-coil-challenge}}

\label{sec:coil}

In this section, we briefly describe a binary classification example using data from the CoIL 2000 Challenge (Putten and Someren 2004); the code for this example, which uses R (and the \href{}{reticulate} package (Ushey, Allaire, and Tang 2021) to interface directly with Python), is given in the appendix.

The data, which we obtained from the R package \href{https://cran.r-project.org/package=kernlab}{kernlab} (Karatzoglou, Smola, and Hornik 2019), consists of 9822 customer records containing 86 variables, including product usage data and socio-demographic data derived from zip area codes. The goal of the challenge was to be able to answer the following question: ``Can you predict who would be interested in buying a caravan insurance policy and give an explanation of why?'' Hence, being able to explain your model's predictions was key to being successful in this challenge.

The winning entry, by Charles Elkan (Elkan 2001), correctly identified 121 caravan policy holders among its 800 top predictions. Our initial EBM (see the R code in the appendix) correctly identified 131. After applying the LASSO, we ended up with a simplified EBM with only 13 terms (quite a reduction from the original 96, counting the intercept)! Moreover, the simplified EBM actually did slightly better and successfully identified 134 policy holders using the top 800 predictions. Woot! Well, technically, the reduced model doesn't provide a fair comparison here since we already used the test data to estimate the LASSO coefficients, which is technically leakage\footnote{In an ideal setting, we'd hopefully have access to a separate and independent sample to use for estimating the LASSO coefficients, and which would not be used in any of the validation or final model comparisons.}.

The results from applying the LASSO are displayed in Figure \ref{fig:coil-code}. The deviance (or twice the log-loss) for the reduced model is 0.420 compared to 0.414 for the original model. Similarly, the AUROC for the reduced model only dropped to 0.728 from the initial 0.755 for the full model. Again, we seem to have found a valuable reduction in complexity from the original model for only a modest cost to the performance (although, we seemed to do better in terms of gain, which the original challenge focused on).

\hypertarget{discussion}{%
\section{Discussion}\label{discussion}}

We have discussed a relatively straightforward approach to simplifying complex EBM models with many terms by reweighting the individual term contributions using coefficients estimated by the LASSO. In high-dimensional problems (i.e., data sets with many variables), this approach can often reduce the number of terms in the EBM model with little sacrifice to model performance, thereby enhancing the glass-box nature of these models while retaining comparable accuracy. Consequently, this method is unlikely to provide much benefit in problems where the number of features is relatively small.

This approach can be applied to more general GA\textsuperscript{2}M models as well. Furthermore, this approach can be generalized in several useful way, but we have not actively explored any of them. A few of our ideas for enhancing this approach are listed below:

\begin{itemize}
\tightlist
\item
  Using the group LASSO (Yuan and Lin 2005) to maintain groups of terms (e.g., making sure that if one term is in the model, then the other terms in the group will also have non-zero coefficients).
\item
  Using post-selection inference (Taylor and Tibshirani 2016; Lee et al. 2016) to obtain \(p\)-values and confidence intervals for the non-zero coefficients.
\item
  Using a version of the LASSO that follows the \emph{hierarchical principle}, as proposed in Bien, Taylor, and Tibshirani (2013). Here, an additional hierarchy constraint is added to ensure that a pairwise interaction only be included if one or both predictors are also included as main effects.
\end{itemize}

\appendix

\section{R code for the CoIL 2000 Challenge example}

\begin{Shaded}
\begin{Highlighting}[]
\CommentTok{\# install.packages(c("glmnet", "kernlab", "reticulate", "treemisc"))}
\FunctionTok{library}\NormalTok{(glmnet)}

\CommentTok{\#}
\CommentTok{\# This assumes that the user has properly setup the \textasciigrave{}reticulate\textasciigrave{} package to use }
\CommentTok{\# an appropriate Python environment with the \textasciigrave{}interpret\textasciigrave{} and \textasciigrave{}sklearn\textasciigrave{} libraries}
\CommentTok{\# available. For details on getting started with \textasciigrave{}reticulate\textasciigrave{} package, visit}
\CommentTok{\# https://rstudio.github.io/reticulate/.}
\CommentTok{\#}

\CommentTok{\# Load required Python libraries}
\NormalTok{interpret }\OtherTok{\textless{}{-}}\NormalTok{ reticulate}\SpecialCharTok{::}\FunctionTok{import}\NormalTok{(}\StringTok{"interpret"}\NormalTok{)}
\NormalTok{skl }\OtherTok{\textless{}{-}}\NormalTok{ reticulate}\SpecialCharTok{::}\FunctionTok{import}\NormalTok{(}\StringTok{"sklearn"}\NormalTok{)}

\CommentTok{\# Split into train/test sets using predefined test set indicator}
\FunctionTok{data}\NormalTok{(ticdata, }\AttributeTok{package =} \StringTok{"kernlab"}\NormalTok{)  }\CommentTok{\# data is in \textasciigrave{}kernlab\textasciigrave{} package}
\NormalTok{trn }\OtherTok{\textless{}{-}}\NormalTok{ ticdata[}\DecValTok{1}\SpecialCharTok{:}\DecValTok{5000}\NormalTok{, ]}
\NormalTok{tst }\OtherTok{\textless{}{-}}\NormalTok{ ticdata[}\SpecialCharTok{{-}}\NormalTok{(}\DecValTok{1}\SpecialCharTok{:}\DecValTok{5000}\NormalTok{), ]}
\NormalTok{X\_trn }\OtherTok{\textless{}{-}} \FunctionTok{subset}\NormalTok{(trn, }\AttributeTok{select =} \SpecialCharTok{{-}}\NormalTok{CARAVAN)  }\CommentTok{\# feature columns}
\NormalTok{X\_tst }\OtherTok{\textless{}{-}} \FunctionTok{subset}\NormalTok{(tst, }\AttributeTok{select =} \SpecialCharTok{{-}}\NormalTok{CARAVAN)}
\NormalTok{y\_trn }\OtherTok{\textless{}{-}} \FunctionTok{ifelse}\NormalTok{(trn}\SpecialCharTok{$}\NormalTok{CARAVAN }\SpecialCharTok{==} \StringTok{"insurance"}\NormalTok{, }\DecValTok{1}\NormalTok{, }\DecValTok{0}\NormalTok{)  }\CommentTok{\# response vector as 0/1}
\NormalTok{y\_tst }\OtherTok{\textless{}{-}} \FunctionTok{ifelse}\NormalTok{(tst}\SpecialCharTok{$}\NormalTok{CARAVAN }\SpecialCharTok{==} \StringTok{"insurance"}\NormalTok{, }\DecValTok{1}\NormalTok{, }\DecValTok{0}\NormalTok{)}

\CommentTok{\# Fit a basic EBM classifier}
\NormalTok{EBC }\OtherTok{\textless{}{-}}\NormalTok{ interpret}\SpecialCharTok{$}\NormalTok{glassbox}\SpecialCharTok{$}\NormalTok{ExplainableBoostingClassifier}
\NormalTok{ebm }\OtherTok{\textless{}{-}} \FunctionTok{EBC}\NormalTok{(}\AttributeTok{inner\_bags =}\NormalTok{ 25L, }\AttributeTok{outer\_bags =}\NormalTok{ 25L)  }\CommentTok{\# need to pass integers}
\NormalTok{ebm}\SpecialCharTok{$}\FunctionTok{fit}\NormalTok{(X\_trn, }\AttributeTok{y =}\NormalTok{ y\_trn)  }\CommentTok{\# fit the model}
\end{Highlighting}
\end{Shaded}

\begin{verbatim}
## ExplainableBoostingClassifier(inner_bags=25, outer_bags=25)
\end{verbatim}

\begin{Shaded}
\begin{Highlighting}[]
\CommentTok{\# How many policy holders would we successfully identify by using the top 800}
\CommentTok{\# predictions? (For comparison with competition winner.)}
\NormalTok{p }\OtherTok{\textless{}{-}}\NormalTok{ ebm}\SpecialCharTok{$}\FunctionTok{predict\_proba}\NormalTok{(X\_tst)[, 2L]  }\CommentTok{\# P(Y=1|X)}
\FunctionTok{sum}\NormalTok{(y\_tst[}\FunctionTok{order}\NormalTok{(p, }\AttributeTok{decreasing =} \ConstantTok{TRUE}\NormalTok{)][}\DecValTok{1}\SpecialCharTok{:}\DecValTok{800}\NormalTok{])}
\end{Highlighting}
\end{Shaded}

\begin{verbatim}
## [1] 131
\end{verbatim}

\begin{Shaded}
\begin{Highlighting}[]
\CommentTok{\# Compute binomial deviance (this is used by \textasciigrave{}glmnet\textasciigrave{} and is equivalent to }
\CommentTok{\# twice the log{-}loss here) and AUROC on the test data}
\NormalTok{(bin\_dev }\OtherTok{\textless{}{-}}\NormalTok{ skl}\SpecialCharTok{$}\NormalTok{metrics}\SpecialCharTok{$}\FunctionTok{log\_loss}\NormalTok{(y\_tst, ebm}\SpecialCharTok{$}\FunctionTok{predict\_proba}\NormalTok{(X\_tst)[, 2L]) }\SpecialCharTok{*} \DecValTok{2}\NormalTok{)}
\end{Highlighting}
\end{Shaded}

\begin{verbatim}
## [1] 0.4138882
\end{verbatim}

\begin{Shaded}
\begin{Highlighting}[]
\NormalTok{(au\_roc }\OtherTok{\textless{}{-}}\NormalTok{ skl}\SpecialCharTok{$}\NormalTok{metrics}\SpecialCharTok{$}\FunctionTok{roc\_auc\_score}\NormalTok{(y\_tst, ebm}\SpecialCharTok{$}\FunctionTok{predict\_proba}\NormalTok{(X\_tst)[, 2L]))}
\end{Highlighting}
\end{Shaded}

\begin{verbatim}
## [1] 0.7559392
\end{verbatim}

\begin{Shaded}
\begin{Highlighting}[]
\CommentTok{\# Print number of terms in the model (not counting intercept)}
\FunctionTok{length}\NormalTok{(ebm}\SpecialCharTok{$}\NormalTok{term\_names\_) }\SpecialCharTok{+} \DecValTok{1}  \CommentTok{\# add 1 to count for intercept}
\end{Highlighting}
\end{Shaded}

\begin{verbatim}
## [1] 96
\end{verbatim}

\begin{Shaded}
\begin{Highlighting}[]
\CommentTok{\# Function to grab matrix of individual term contributions (no intercept)}
\NormalTok{predict\_terms }\OtherTok{\textless{}{-}} \ControlFlowTok{function}\NormalTok{(object, newdata) \{}
\NormalTok{  contrib }\OtherTok{\textless{}{-}}\NormalTok{ object}\SpecialCharTok{$}\FunctionTok{predict\_and\_contrib}\NormalTok{(newdata)[[2L]]  }\CommentTok{\# grab second component}
  \FunctionTok{colnames}\NormalTok{(contrib) }\OtherTok{\textless{}{-}}\NormalTok{ ebm}\SpecialCharTok{$}\NormalTok{term\_names\_  }\CommentTok{\# add column names}
\NormalTok{  contrib  }\CommentTok{\# Note: rowSum(contrib) + ebm$intercept\_ = ebm$predict(newdata)}
\NormalTok{\}}

\CommentTok{\# Compute matrix of individual term contributions for train and test sets}
\NormalTok{contrib\_trn }\OtherTok{\textless{}{-}} \FunctionTok{predict\_terms}\NormalTok{(ebm, }\AttributeTok{newdata =}\NormalTok{ X\_trn)}
\NormalTok{contrib\_tst }\OtherTok{\textless{}{-}} \FunctionTok{predict\_terms}\NormalTok{(ebm, }\AttributeTok{newdata =}\NormalTok{ X\_tst)}

\CommentTok{\# Sanity check that term contributions are additive to prediction on link scale}
\FunctionTok{head}\NormalTok{(}\FunctionTok{cbind}\NormalTok{(}
  \StringTok{"predict\_proba"} \OtherTok{=}\NormalTok{ ebm}\SpecialCharTok{$}\FunctionTok{predict\_proba}\NormalTok{(X\_tst)[, 2L],}
  \CommentTok{\# Could also use \textasciigrave{}boot::inv.logit()\textasciigrave{} instead of \textasciigrave{}plogis()\textasciigrave{} for inverse logit}
  \StringTok{"added\_terms"} \OtherTok{=} \FunctionTok{plogis}\NormalTok{(}\FunctionTok{rowSums}\NormalTok{(contrib\_tst) }\SpecialCharTok{+} \FunctionTok{c}\NormalTok{(ebm}\SpecialCharTok{$}\NormalTok{intercept\_))}
\NormalTok{))}
\end{Highlighting}
\end{Shaded}

\begin{verbatim}
##      predict_proba added_terms
## [1,]   0.013915601 0.013915601
## [2,]   0.013351104 0.013351104
## [3,]   0.006006496 0.006006496
## [4,]   0.081980003 0.081980003
## [5,]   0.019211908 0.019211908
## [6,]   0.035693249 0.035693249
\end{verbatim}

\begin{Shaded}
\begin{Highlighting}[]
\CommentTok{\# Fit the LASSO regularization path using the term contributions as inputs}
\NormalTok{ebm\_lasso }\OtherTok{\textless{}{-}} \FunctionTok{glmnet}\NormalTok{(}
  \AttributeTok{x =}\NormalTok{ contrib\_trn,      }\CommentTok{\# individual term contributions}
  \AttributeTok{y =}\NormalTok{ y\_trn,            }\CommentTok{\# original response variable}
  \AttributeTok{lower.limits =} \DecValTok{0}\NormalTok{,     }\CommentTok{\# coefficients should be strictly positive}
  \AttributeTok{standardize =} \ConstantTok{FALSE}\NormalTok{,  }\CommentTok{\# no need to standardize}
  \AttributeTok{family =} \StringTok{"binomial"}   \CommentTok{\# logistic regression}
\NormalTok{)}

\CommentTok{\# Assess performance of fit using an independent test set}
\NormalTok{perf }\OtherTok{\textless{}{-}} \FunctionTok{assess.glmnet}\NormalTok{(}
  \AttributeTok{object =}\NormalTok{ ebm\_lasso,  }\CommentTok{\# fitted LASSO model}
  \AttributeTok{newx =}\NormalTok{ contrib\_tst,  }\CommentTok{\# test set contributions}
  \AttributeTok{newy =}\NormalTok{ y\_tst,        }\CommentTok{\# same response variable (test set)}
  \AttributeTok{family =} \StringTok{"binomial"}  \CommentTok{\# for logit link (i.e., logistic regression)}
\NormalTok{)}
\NormalTok{perf }\OtherTok{\textless{}{-}} \FunctionTok{do.call}\NormalTok{(cbind, }\AttributeTok{args =}\NormalTok{ perf)  }\CommentTok{\# bind results into matrix}

\CommentTok{\# Data frame of results (one row for each value of lambda)}
\NormalTok{ebm\_lasso\_results }\OtherTok{\textless{}{-}} \FunctionTok{as.data.frame}\NormalTok{(}\FunctionTok{cbind}\NormalTok{(}
  \StringTok{"num\_terms"} \OtherTok{=}\NormalTok{ ebm\_lasso}\SpecialCharTok{$}\NormalTok{df,  }\CommentTok{\# number of non{-}zero coefficients for each lambda}
\NormalTok{  perf,  }\CommentTok{\# performance metrics (i.e., deviance, mse, auc, etc.)}
  \StringTok{"lambda"} \OtherTok{=}\NormalTok{ ebm\_lasso}\SpecialCharTok{$}\NormalTok{lambda}
\NormalTok{))}

\CommentTok{\# Sort in ascending order of number of trees}
\FunctionTok{head}\NormalTok{(ebm\_lasso\_results }\OtherTok{\textless{}{-}} 
\NormalTok{       ebm\_lasso\_results[}\FunctionTok{order}\NormalTok{(ebm\_lasso\_results}\SpecialCharTok{$}\NormalTok{num\_terms), ])}
\end{Highlighting}
\end{Shaded}

\begin{verbatim}
##    num_terms  deviance      class       auc       mse       mae      lambda
## s0         0 0.4593780 0.06097055 0.5000000 0.1145195 0.2244984 0.007699324
## s1         1 0.4548637 0.06097055 0.6617197 0.1140162 0.2241129 0.007015337
## s2         1 0.4512166 0.06097055 0.6617197 0.1136000 0.2237568 0.006392114
## s3         1 0.4482546 0.06097055 0.6617197 0.1132565 0.2234288 0.005824256
## s4         1 0.4458408 0.06097055 0.6617197 0.1129732 0.2231273 0.005306845
## s5         1 0.4438697 0.06097055 0.6617197 0.1127400 0.2228505 0.004835400
\end{verbatim}

\begin{Shaded}
\begin{Highlighting}[]
\CommentTok{\# Print results corresponding to smallest test deviance (or half the log{-}loss)}
\NormalTok{ebm\_lasso\_results[}\FunctionTok{which.min}\NormalTok{(ebm\_lasso\_results}\SpecialCharTok{$}\NormalTok{deviance), ]}
\end{Highlighting}
\end{Shaded}

\begin{verbatim}
##     num_terms deviance      class       auc       mse       mae       lambda
## s26        12 0.420382 0.06138532 0.7275167 0.1094124 0.2143992 0.0006854054
\end{verbatim}

\begin{Shaded}
\begin{Highlighting}[]
\CommentTok{\# Grab "best" lambda according to test deviance}
\NormalTok{(lambda }\OtherTok{\textless{}{-}}\NormalTok{ ebm\_lasso\_results[}\FunctionTok{which.min}\NormalTok{(ebm\_lasso\_results}\SpecialCharTok{$}\NormalTok{deviance), }\StringTok{"lambda"}\NormalTok{])}
\end{Highlighting}
\end{Shaded}

\begin{verbatim}
## [1] 0.0006854054
\end{verbatim}

\begin{Shaded}
\begin{Highlighting}[]
\NormalTok{p }\OtherTok{\textless{}{-}} \FunctionTok{predict}\NormalTok{(ebm\_lasso, }\AttributeTok{newx =}\NormalTok{ contrib\_tst, }\AttributeTok{s =}\NormalTok{ lambda, }\AttributeTok{type =} \StringTok{"response"}\NormalTok{)[, }\DecValTok{1}\NormalTok{]}
\FunctionTok{sum}\NormalTok{(y\_tst[}\FunctionTok{order}\NormalTok{(p, }\AttributeTok{decreasing =} \ConstantTok{TRUE}\NormalTok{)][}\DecValTok{1}\SpecialCharTok{:}\DecValTok{800}\NormalTok{])}
\end{Highlighting}
\end{Shaded}

\begin{verbatim}
## [1] 134
\end{verbatim}

\begin{Shaded}
\begin{Highlighting}[]
\CommentTok{\# Plot results }
\FunctionTok{par}\NormalTok{(}
  \AttributeTok{mfrow =} \FunctionTok{c}\NormalTok{(}\DecValTok{1}\NormalTok{, }\DecValTok{2}\NormalTok{),}
  \AttributeTok{mar =} \FunctionTok{c}\NormalTok{(}\DecValTok{4}\NormalTok{, }\DecValTok{4}\NormalTok{, }\FloatTok{0.1}\NormalTok{, }\FloatTok{0.1}\NormalTok{), }
  \AttributeTok{cex.lab =} \FloatTok{0.95}\NormalTok{, }
  \AttributeTok{cex.axis =} \FloatTok{0.8}\NormalTok{,  }\CommentTok{\# was 0.9}
  \AttributeTok{mgp =} \FunctionTok{c}\NormalTok{(}\DecValTok{2}\NormalTok{, }\FloatTok{0.7}\NormalTok{, }\DecValTok{0}\NormalTok{), }
  \AttributeTok{tcl =} \SpecialCharTok{{-}}\FloatTok{0.3}\NormalTok{, }
  \AttributeTok{las =} \DecValTok{1}
\NormalTok{)}
\FunctionTok{palette}\NormalTok{(}\StringTok{"Okabe{-}Ito"}\NormalTok{)}
\FunctionTok{plot}\NormalTok{(ebm\_lasso, }\AttributeTok{xvar =} \StringTok{"lambda"}\NormalTok{, }\AttributeTok{col =} \FunctionTok{adjustcolor}\NormalTok{(}\DecValTok{3}\NormalTok{, }\AttributeTok{alpha.f =} \FloatTok{0.3}\NormalTok{))}
\FunctionTok{abline}\NormalTok{(}\AttributeTok{v =} \FunctionTok{log}\NormalTok{(lambda), }\AttributeTok{lty =} \DecValTok{2}\NormalTok{, }\AttributeTok{col =} \DecValTok{1}\NormalTok{)}
\NormalTok{ylim }\OtherTok{\textless{}{-}} \FunctionTok{range}\NormalTok{(}\FunctionTok{c}\NormalTok{(}\FloatTok{0.4}\NormalTok{, bin\_dev, perf[, }\StringTok{"deviance"}\NormalTok{]))}
\FunctionTok{plot}\NormalTok{(ebm\_lasso\_results[, }\FunctionTok{c}\NormalTok{(}\StringTok{"num\_terms"}\NormalTok{, }\StringTok{"deviance"}\NormalTok{)], }\AttributeTok{type =} \StringTok{"l"}\NormalTok{, }\AttributeTok{las =} \DecValTok{1}\NormalTok{,}
     \AttributeTok{xlab =} \StringTok{"Number of terms"}\NormalTok{, }\AttributeTok{ylab =} \StringTok{"Test deviance"}\NormalTok{, }\AttributeTok{ylim =}\NormalTok{ ylim)}
\FunctionTok{abline}\NormalTok{(}\AttributeTok{h =} \FunctionTok{min}\NormalTok{(ebm\_lasso\_results}\SpecialCharTok{$}\NormalTok{deviance), }\AttributeTok{col =} \DecValTok{1}\NormalTok{, }\AttributeTok{lty =} \DecValTok{2}\NormalTok{)}
\FunctionTok{abline}\NormalTok{(}\AttributeTok{h =}\NormalTok{ bin\_dev, }\AttributeTok{lty =} \DecValTok{2}\NormalTok{, }\AttributeTok{col =} \DecValTok{2}\NormalTok{)}
\FunctionTok{legend}\NormalTok{(}\StringTok{"topleft"}\NormalTok{, }\AttributeTok{inset =} \FloatTok{0.01}\NormalTok{, }\AttributeTok{bty =} \StringTok{"n"}\NormalTok{, }\AttributeTok{col =} \FunctionTok{c}\NormalTok{(}\DecValTok{2}\NormalTok{, }\DecValTok{1}\NormalTok{), }\AttributeTok{lty =} \DecValTok{2}\NormalTok{,}
       \AttributeTok{legend =} \FunctionTok{c}\NormalTok{(}\StringTok{"Original EBM (96 terms)"}\NormalTok{, }\StringTok{"Post LASSO (13 terms)"}\NormalTok{), }\AttributeTok{cex =} \FloatTok{0.8}\NormalTok{)}
\FunctionTok{palette}\NormalTok{(}\StringTok{"default"}\NormalTok{)}
\end{Highlighting}
\end{Shaded}

\begin{figure}[!htb]

{\centering \includegraphics[width=1\linewidth]{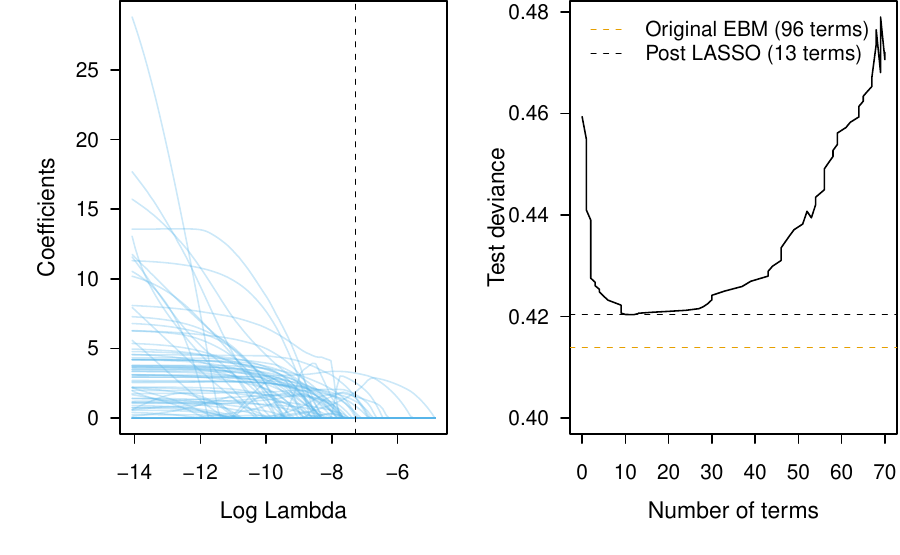} 

}

\caption{Results for the CoIL 2000 Challenge example. Left: LASSO coefficient path showing the estimated coefficient for each term as a function of the penalty parameter $\lambda$ (log scale); the vertical dashed line shows the optimal value of $\lambda$ based on minimizing binomial deviance for the test set. Right: Number of terms with non-zero coefficients as a function of binomial deviance for the test data.}\label{fig:coil-code}
\end{figure}

\newpage

\hypertarget{references}{%
\section*{References}\label{references}}
\addcontentsline{toc}{section}{References}

\hypertarget{refs}{}
\begin{CSLReferences}{1}{0}
\leavevmode\vadjust pre{\hypertarget{ref-bien-2013-lasso}{}}%
Bien, Jacob, Jonathan Taylor, and Robert Tibshirani. 2013. {``{A lasso for hierarchical interactions}.''} \emph{The Annals of Statistics} 41 (3): 1111--41. \url{https://doi.org/10.1214/13-AOS1096}.

\leavevmode\vadjust pre{\hypertarget{ref-breiman-2001-rf}{}}%
Breiman, Leo. 2001. {``Random Forests.''} \emph{Machine Learning} 45 (1): 5--32. \url{https://doi.org/10.1023/A:1010933404324}.

\leavevmode\vadjust pre{\hypertarget{ref-efron-2016-computer}{}}%
Efron, Bradley, and Trevor Hastie. 2016. \emph{Computer Age Statistical Inference: Algorithms, Evidence, and Data Science}. Institute of Mathematical Statistics Monographs. Cambridge University Press. \url{https://doi.org/10.1017/CBO9781316576533}.

\leavevmode\vadjust pre{\hypertarget{ref-elkans-2001-magical}{}}%
Elkan, Charles. 2001. {``Magical Thinking in Data Mining: Lessons from CoIL Challenge 2000.''} In \emph{Proceedings of the Seventh ACM SIGKDD International Conference on Knowledge Discovery and Data Mining}, 426--31. KDD '01. New York, NY, USA: Association for Computing Machinery. \url{https://doi.org/10.1145/502512.502576}.

\leavevmode\vadjust pre{\hypertarget{ref-friedman-2001-greedy}{}}%
Friedman, Jerome H. 2001. {``Greedy Function Approximation: A Gradient Boosting Machine.''} \emph{The Annals of Statistics} 29 (5): 1189--1232. \url{https://doi.org/10.1214/aos/1013203451}.

\leavevmode\vadjust pre{\hypertarget{ref-friedman-2002-stochastic}{}}%
---------. 2002. {``Stochastic Gradient Boosting.''} \emph{Computational Statistics \& Data Analysis} 38 (4): 367--78. https://doi.org/\url{https://doi.org/10.1016/S0167-9473(01)00065-2}.

\leavevmode\vadjust pre{\hypertarget{ref-friedman-2003-isle}{}}%
Friedman, Jerome H., and Bogdan E. Popescu. 2003. {``Importance Sampled Learning Ensembles.''} Technical Report. Stanford University, Department of Statistics. \url{https://statweb.stanford.edu/~jhf/ftp/isle.pdf}.

\leavevmode\vadjust pre{\hypertarget{ref-R-glmnet}{}}%
Friedman, Jerome, Trevor Hastie, Rob Tibshirani, Balasubramanian Narasimhan, Kenneth Tay, and Noah Simon. 2021. \emph{Glmnet: Lasso and Elastic-Net Regularized Generalized Linear Models}. \url{https://CRAN.R-project.org/package=glmnet}.

\leavevmode\vadjust pre{\hypertarget{ref-hastie-2009-elements}{}}%
Hastie, Trevor, Robert. Tibshirani, and Jerome Friedman. 2009. \emph{The Elements of Statistical Learning: Data Mining, Inference, and Prediction, Second Edition}. Springer Series in Statistics. Springer-Verlag. \url{https://web.stanford.edu/~hastie/ElemStatLearn/}.

\leavevmode\vadjust pre{\hypertarget{ref-kapoor-2023-leakage}{}}%
Kapoor, Sayash, and Arvind Narayanan. 2023. {``Leakage and the Reproducibility Crisis in Machine-Learning-Based Science.''} \emph{Patterns}, August. \url{https://doi.org/10.1016/j.patter.2023.100804}.

\leavevmode\vadjust pre{\hypertarget{ref-R-kernlab}{}}%
Karatzoglou, Alexandros, Alex Smola, and Kurt Hornik. 2019. \emph{Kernlab: Kernel-Based Machine Learning Lab}. \url{https://CRAN.R-project.org/package=kernlab}.

\leavevmode\vadjust pre{\hypertarget{ref-kuffner-2015-dream}{}}%
Küffner, Robert, Neta Zach, Raquel Norel, Johann Hawe, David Schoenfeld, Liuxia Wang, Guang Li, et al. 2015. {``Crowdsourced Analysis of Clinical Trial Data to Predict Amyotrophic Lateral Sclerosis Progression.''} \emph{Nature Biotechnology} 33 (1): 51--57. \url{https://doi.org/10.1038/nbt.3051}.

\leavevmode\vadjust pre{\hypertarget{ref-lee-2016-exact}{}}%
Lee, Jason D., Dennis L. Sun, Yuekai Sun, and Jonathan E. Taylor. 2016. {``Exact Post-Selection Inference, with Application to the Lasso.''} \emph{The Annals of Statistics} 44 (3). \url{https://doi.org/10.1214/15-aos1371}.

\leavevmode\vadjust pre{\hypertarget{ref-lou-2013-accurate}{}}%
Lou, Yin, Rich Caruana, Giles Hooker, and Johannes Gehrke. 2013. {``Accurate Intelligible Models with Pairwise Interactions.''} In \emph{KDD'13, August 11--14, 2013, Chicago, Illinois, USA}, KDD'13, August 11--14, 2013, Chicago, Illinois, USA. ACM. \url{https://www.microsoft.com/en-us/research/publication/accurate-intelligible-models-pairwise-interactions/}.

\leavevmode\vadjust pre{\hypertarget{ref-nori-2019-interpretml}{}}%
Nori, Harsha, Samuel Jenkins, Paul Koch, and Rich Caruana. 2019. {``InterpretML: A Unified Framework for Machine Learning Interpretability.''} \emph{arXiv Preprint arXiv:1909.09223}.

\leavevmode\vadjust pre{\hypertarget{ref-putten-2004-coil}{}}%
Putten, Peter van der, and Maarten van Someren. 2004. {``A Bias-Variance Analysis of a Real World Learning Problem: The CoIL Challenge 2000.''} \emph{Machine Learning} 57 (1): 177--95. \url{https://doi.org/10.1023/B:MACH.0000035476.95130.99}.

\leavevmode\vadjust pre{\hypertarget{ref-taylor-2016-post}{}}%
Taylor, Jonathan, and Robert Tibshirani. 2016. {``Post-Selection Inference for L1-Penalized Likelihood Models.''} \url{https://arxiv.org/abs/1602.07358}.

\leavevmode\vadjust pre{\hypertarget{ref-tibshirani-1996-lasso}{}}%
Tibshirani, Robert. 1996. {``Regression Shrinkage and Selection via the Lasso.''} \emph{Journal of the Royal Statistical Society. Series B (Methodological)} 58 (1): 267--88.

\leavevmode\vadjust pre{\hypertarget{ref-R-reticulate}{}}%
Ushey, Kevin, JJ Allaire, and Yuan Tang. 2021. \emph{Reticulate: Interface to Python}. \url{https://github.com/rstudio/reticulate}.

\leavevmode\vadjust pre{\hypertarget{ref-wick-2020-cyclic}{}}%
Wick, Felix, Ulrich Kerzel, and Michael Feindt. 2020. {``Cyclic Boosting - an Explainable Supervised Machine Learning Algorithm.''} \emph{CoRR} abs/2002.03425. \url{https://arxiv.org/abs/2002.03425}.

\leavevmode\vadjust pre{\hypertarget{ref-wu-2014-nonnegative}{}}%
Wu, Lan, Yuehan Yang, and Hanzhong Liu. 2014. {``Nonnegative-Lasso and Application in Index Tracking.''} \emph{Computational Statistics \& Data Analysis} 70: 116--26. https://doi.org/\url{https://doi.org/10.1016/j.csda.2013.08.012}.

\leavevmode\vadjust pre{\hypertarget{ref-yuan-2005-group}{}}%
Yuan, Ming, and Yi Lin. 2005. {``{Model Selection and Estimation in Regression with Grouped Variables}.''} \emph{Journal of the Royal Statistical Society Series B: Statistical Methodology} 68 (1): 49--67. \url{https://doi.org/10.1111/j.1467-9868.2005.00532.x}.

\end{CSLReferences}

\bibliographystyle{unsrt}
\bibliography{references.bib}

\end{document}